\documentclass[
]{ceurart}

\usepackage{listings}
\usepackage{amsmath,amssymb,amsfonts}
\usepackage{algorithmic}
\usepackage{graphicx}
\usepackage{textcomp}
\usepackage{xcolor}
\usepackage{subcaption}
\usepackage{algorithm}
\usepackage{url}
\usepackage{cleveref} 
\usepackage{enumitem}

\begin{document}

\copyrightyear{2026}
\copyrightclause{Copyright for this paper by its authors.
  Use permitted under Creative Commons License Attribution 4.0
  International (CC BY 4.0).}

\conference{ICRA'26: ROSE International Workshop on Robotics Software Engineering,
  June 01, 2026, Vienna, Austria}

\title{CrossMaps: Confidence-Aware Open-Vocabulary Semantic Mapping for Rover Navigation}

%\tnotetext[1]{You can use this document as the template for preparing your
%  publication. We recommend using the latest version of the ceurart style.}


\author[]{Jan-Niklas Klein}[%
email=jan-niklas.klein@student.hpi.uni-potsdam.de
]

\author[]{Sona Ghahremani}[%
orcid=0000-0003-0697-9195,
email=sona.ghahremani@hpi.de,
%url=https://hpi.de/en/giese/people/sona-ghahremani.html,
]
\cormark[1]

\author[]{Christian Medeiros Adriano}[%
orcid=0000-0003-2588-9937,
email=christian.adriano@hpi.de,
%url=https://hpi.de/en/giese/people/christian-medeiros-adriano.html,
]
\cormark[1]
\author[]{Holger Giese}[%
orcid=0000-0002-4723-730X,
email=holger.giese@hpi.de,
%url=https://hpi.de/en/giese/people/prof-dr-holger-giese.html,
]
\address[]{Hasso Plattner Institute for Digital Engineering, Potsdam, Germany.}
\cortext[1]{Corresponding author.}

\begin{abstract}
%Rovers rely on perception to maintain spatial maps that encode both objects and sensor quality (e.g., range reliability, lighting artifacts, data density), which drives decisions about data fusion, embedding updates, and navigation re-planning under partial observability. To study these coupled perception–navigation decisions, we present CrossMaps, a real-time, confidence-aware open-vocabulary semantic mapping pipeline for mobile rovers that constructs language-queryable maps from RGB-D data. Building on VLMaps-style methods, CrossMaps integrates multi-scale CLIP embeddings with confidence-aware fusion and a dual-memory architecture comprising Short-Term Memory (STM) and Long-Term Memory (LTM). The STM consolidates noisy visual observations using geometric, semantic, and temporal confidence cues with consistency gating and multi-view tracking, while the sufficiently confident and coherent cells are promoted to the LTM as persistent semantic landmarks. Designed for deployment with a Jetson Orin-powered UGV alongside SLAM, CrossMaps runs in real time and produces semantic heatmaps that can be queried with natural language to guide rover navigation in evolving environments.
Rovers rely on perception to maintain spatial maps that encode both objects and sensor quality (e.g., range reliability, lighting artifacts, data density), guiding data fusion, embedding updates, and navigation under partial observability. To study these coupled perception–navigation processes, we present CrossMaps, a real-time confidence-aware open-vocabulary semantic mapping pipeline that constructs language-queryable maps from RGB-D data. Building on VLMaps-style approaches, CrossMaps integrates multi-scale CLIP embeddings with confidence-aware fusion and a dual-memory architecture consisting of Short-Term Memory (STM) and Long-Term Memory (LTM). The STM aggregates noisy visual observations using geometric, semantic, and temporal confidence cues, while confident and coherent cells are promoted to the LTM as persistent semantic landmarks. Designed for deployment with a Jetson Orin-powered UGV alongside SLAM, CrossMaps runs in real time and produces semantic heatmaps that can be queried with natural language to guide rover navigation.
\end{abstract}


\maketitle
\vspace{-3mm}
\section{Introduction}

\textbf{Context and Limitations} -- Autonomous rovers in unstructured environments must continuously build representations of their surroundings to navigate safely and achieve mission goals, a task that is especially challenging for planetary rovers due to partial observability, sensor noise, and dynamic conditions. Traditional navigation systems rely mainly on SLAM-based geometric maps~\cite{Labbe2019RTAB,Campos2021ORB}, which capture spatial structure but lack semantic understanding of the environment. To address this semantic gap, recent advances in VLMs (Vision Language Models) have enabled open-vocabulary semantic mapping, allowing rovers to associate spatial locations with natural language concepts using methods such as VLMaps~\cite{Huang2023VLMaps}, which augment geometric maps with CLIP-based embeddings~\cite{radford2021learning} for text-queryable navigation beyond predefined object classes. However, deploying such systems on mobile rovers remains challenging: observations are noisy and viewpoint-dependent, embeddings vary across viewpoints and lighting, objects may appear or disappear, and the map must remain stable over long horizons while still adapting to new evidence, requiring mechanisms that carefully balance plasticity with robustness.

%-- Recent advances in VLMs have enabled open-vocabulary semantic mapping, enabling rovers to associate spatial locations with natural language concepts. Methods such as Visual Language Maps (VLMaps) augment spatial maps with semantic embeddings derived from models such as CLIP \cite{radford2021learning}, allowing rovers to query maps using arbitrary text descriptions. These approaches significantly extend the capabilities of rover navigation systems beyond predefined object classes.
%\textbf{Current Limitations} -- However, deploying such systems on mobile rovers introduces several practical challenges. Rovers collect observations as they move, yielding noisy, viewpoint-dependent visual inputs. Semantic embeddings derived from images may vary across viewpoints or lighting conditions, and dynamic objects may appear or disappear over time. Furthermore, a semantic map must remain stable over long time horizons while still adapting to new observations. These challenges require mechanisms that balance plasticity, the ability to integrate new information, with robustness against noise and transient observations. 
\noindent\textbf{Research Problems} -- We investigate the transfer of perception knowledge along three main axes:
%\begin{enumerate}

    \textbf{RP.1} - \textit{How to capture and represent perception knowledge?} We focus on perception knowledge with both qualitative (e.g., semantic information) and quantitative (e.g., distance) properties. %Our approach is to experiment with short- and long-term memory structures that aggregate confidence (uncertainty), coherence (agreement), and semantic information.
    
    \textbf{RP.2} - \textit{How to pre-process, clean, and rank perceptual hypotheses, given their spatial and temporal information as well as their inherent uncertainty?} The primary source of perception knowledge is sensor data, which can be noisy, incomplete, and biased due to imperfections that may depend on the rover’s actions (e.g., viewpoint, occlusions) or not (e.g., lighting, weather).%, we study a framework that combines i.i.d. observation uncertainty (confidence), agreement across viewpoints (coherence), and scene dynamics (temporal decay).
    
    \textbf{RP.3} - \textit{How can we maintain, merge, and share perception knowledge across multiple agents with different downstream tasks while preserving semantic consistency?} We study update and fusion rules that keep shared perception maps semantically consistent while exposing task-relevant subsets for agents with different intentions (e.g., exploration, inspection).
 %, enabling reliable multi-agent knowledge transfer.
%\end{enumerate}

\noindent\textbf{Approach} -- We introduce \emph{CrossMaps}, a confidence-aware open-vocabulary semantic mapping pipeline designed for mobile rover navigation. CrossMaps extends the VLMaps paradigm by introducing a dual-memory representation and confidence-aware data fusion mechanisms that improve robustness during online mapping. The pipeline maintains a \textit{Short-Term Memory (STM)} that aggregates recent observations while filtering noise through temporal decay and consistency checks, and a \textit{Long-Term Memory (LTM)} that stores reliable semantic landmarks derived from confident observations.

\noindent\textbf{Contributions} -- %(i) A confidence-aware open-vocabulary semantic mapping framework for rover navigation. 
%(i) A confidence-aware open-vocabulary semantic mapping framework that estimates point-wise confidence for CLIP features based on geometric, semantic, and visibility cues before projection into the map,
(i) a confidence-aware open-vocabulary semantic mapping framework that assigns point- and cell-level confidence to CLIP-based observations using geometric, semantic, and temporal cues during map fusion, 
%(ii) A dual-memory architecture separating transient semantic observations from persistent landmarks. 
%(ii) A dual-memory semantic map representation that aggregates these points into cell-wise confidence and coherence in a short-term grid map, and promotes only sufficiently confident, coherent, and multi-view–supported cells into a persistent long-term landmark map.
(ii) a dual-memory semantic map representation that aggregates observations into cell-wise embeddings, confidence, and coherence in a short-term
grid map, and promotes only sufficiently confident, coherent, and multi-view–supported cells into a persistent long-term landmark map, 
%(iii) A real-time semantic mapping pipeline integrating multi-scale CLIP embeddings with SLAM-based spatial mapping. 
%(iii) A real-time semantic mapping pipeline that fuses multi-scale, tile-level CLIP embeddings with SLAM-based spatial mapping to build language-queryable 3D maps under rover motion.
%(iv) A confidence-based promotion mechanism ensuring that only reliable observations become persistent semantic map elements.
(iii) a real-time semantic mapping pipeline that
fuses multi-scale, tile-level CLIP embeddings with SLAM-based spatial alignment to build language-queryable 3D maps under rover motion, and finally, 
(iv) a confidence- and coherence-based promotion and querying mechanism that uses coherence (intra-cell agreement in embedding space) to shape heatmaps and cell selection, while using cell-wise confidence to gate long-term promotion and top-1 navigation targets.

\section{Related Work}

Research on rover navigation mapping has progressed from geometric SLAM toward semantically enriched, language-aware representations. Classical feature-based and graph-based SLAM systems such as ORB-SLAM3~\cite{Campos2021ORB} and RTAB-Map~\cite{Labbe2019RTAB} focus on accurate localization and geometric reconstruction using feature tracking, bundle adjustment, loop closure, and multi-modal inputs, but they do not provide language-aligned semantics suitable for open-vocabulary interaction. More recent dense and neural SLAM approaches like NICE-SLAM~\cite{Zhu2022NICESLAM} improve geometric fidelity using neural implicit representations, but lack open-vocabulary semantic reasoning for navigation.

Semantic SLAM augments geometric maps with semantic categories or object instances, typically relying on closed-set detectors such as YOLO~\cite{Redmon2016YOLO} or Mask R-CNN~\cite{He2017MaskRCNN} sometimes combined with general segmentation models like SAM~\cite{Kirillov2023SAM}. While effective in structured settings, these systems are constrained by predefined semantic vocabularies and struggle to generalize in open-world environments, limiting their applicability to open-vocabulary rover missions~\cite{Redmon2016YOLO,He2017MaskRCNN,Kirillov2023SAM}.

Open-vocabulary 3D scene understanding leverages vision–language models to overcome fixed label sets. Methods such as OpenScene~\cite{Peng2023OpenScene} and OpenOcc~\cite{Jiang2024OpenOcc} project CLIP-based embeddings into 3D reconstructions or volumetric occupancy grids, enabling zero-shot recognition beyond predefined object categories. However, these approaches are designed for offline scene processing and do not address dynamic, resource-constrained navigation with continuous map updates.

Recent work integrates vision–language models directly into online mapping. VLMaps~\cite{Huang2023VLMaps}, LERF~\cite{Kerr2023LERF}, Open-Fusion~\cite{Yamazaki2023OpenFusion} and subsequent systems construct spatially indexed or object-centric maps that can be queried with natural language, and more recent frameworks like FindAnything~\cite{Laina2026FindAnything} and DualMap~\cite{jiang2025dualmap} begin to address online operation and dynamic environments. These methods demonstrate the feasibility of open-vocabulary mapping but generally fuse all evidence into a single representation with limited treatment of uncertainty. %Conversely, CrossMaps introduces confidence-aware fusion and an explicit dual-memory (STM/LTM) architecture that separates noisy short-term evidence from persistent semantic landmarks for rover navigation.

%\small
%\begin{table}[t]
%\centering
%\caption{Comparison of open-vocabulary semantic mapping approaches (reversed).}
%\label{tab:related_methods_reversed}
%\begin{tabular}{lllllll}
%\hline
%\textbf{Method} & \textbf{Online} & \textbf{Loop} & %\textbf{Geometry } & \textbf{Environ.} & \textbf{Confid.} %& \textbf{Memory}  \\
%\hline
%VLMaps & Partial & No & Occupancy + CLIP  & Static & No & Single map \\
%OpenFusion & Yes & Yes & TSDF & Static & Partial & Single map \\
%OpenScene & No & No & Point cloud, voxel CLIP & Static & No & Offline reconstr. \\
%LERF & No & No & NeRF & Static & No & Neural field \\
%OVO  & Partial & No & TSDF + SLAM & Partial & Partial & Partial memory  \\
%FindAnything & Yes & Yes & Object-centric map & Partial & %Partial & Object submaps \\
%DualMap & Yes & Yes & Dynam. seman. map & Dynam.& Partial & Dual represent.  \\
%\textbf{CrossMaps} & Yes & Yes & Grid + seman. embed. & Dynam. & Yes & STM + LTM \\
%\hline
%\end{tabular}
%\end{table}
%\normalsize 
%In summary, Table~\ref{tab:related_methods_reversed} summarizes the main differences between existing open-vocabulary mapping systems and our proposed CrossMaps framework. 

%Conversely, CrossMaps integrates confidence-aware fusion and a dual-memory representation that separates short-term observations from persistent semantic landmarks.

\section{Approach - The CrossMaps Pipeline}\label{sec:approach}

\begin{figure}[b]
\begin{centering}
%\vspace{-2mm}
  \includegraphics[width=0.8\linewidth]{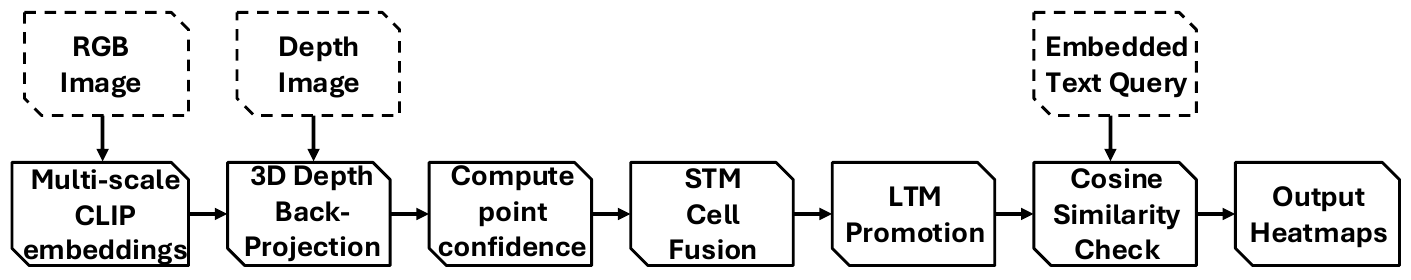}
  \caption{CrossMaps pipeline }
  \label{fig:pipeline}
  \end{centering}
%\vspace{-3mm}
\end{figure}

\textbf{Overview} -- We extended DualMaps~\cite{jiang2025dualmap} and VLMaps~\cite{Huang2023VLMaps} by constructing an open-vocabulary semantic map that integrates vision-language embeddings with spatial mapping and confidence-aware memory management. The pipeline processes a stream of RGB-D observations during rover navigation and, given an RGB image $I_t$, depth image $D_t$, and rover (camera) viewpoint $T_t$ estimated by SLAM at time $t$, extracts CLIP-based semantic embeddings, back-projects them into a 3D spatial grid map, and fuses them into short-term memory using confidence-aware updates. It maintains both an STM that aggregates recent observations and an LTM that stores persistent semantic landmarks. 
%
%The four stages of the pipeline (\Cref{fig:pipeline}) consist of: (1) extraction of semantic embeddings from RGB observations, (2) projection of embeddings into spatial map cells using depth and viewpoint information, (3) confidence-aware fusion of observations in STM, and (4) promotion of semantic cells into LTM—while operating alongside a SLAM pipeline that continuously updates the semantic map as the rover moves (see Algorithm~\ref{alg:crossmaps}).
%
The pipeline (\Cref{fig:pipeline}) consists of embedding extraction, spatial projection into spatial map cells using depth and viewpoint information, confidence-aware fusion of observations in STM, and STM-to-LTM promotion (see Algorithm~\ref{alg:crossmaps}).

%(Contribution 3
\begin{figure}[t]
  \centering
  \begin{subfigure}[t]{0.3\linewidth}
    \centering
    \includegraphics[width=\linewidth]{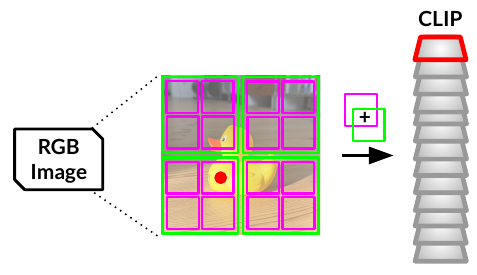}
    \caption{Tile-level coarse~(green) and fine~(pink) embeddings}
    \label{fig:CLIP}
  \end{subfigure}
  \hfill
  \begin{subfigure}[t]{0.2\linewidth}
    \centering
 \includegraphics[width=\linewidth]{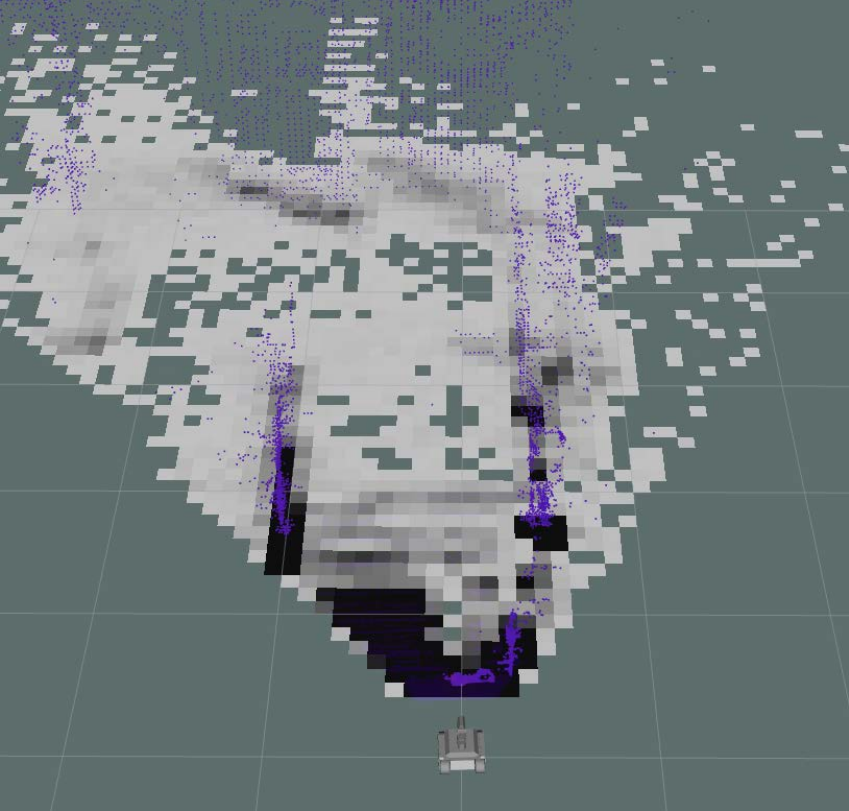}
    \caption{Cell-confidence semantic map}
    \label{fig:pointConf}
  \end{subfigure}
  \hfill
  \begin{subfigure}[t]{0.4\linewidth}
    \centering
\includegraphics[width=\linewidth]{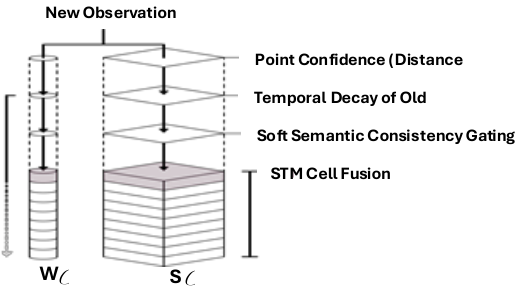}
    \caption{STM cell fusion}
    \label{fig:Cell}  
  \end{subfigure}
  \caption{Visualization of transformations across data representations}
\end{figure}

\paragraph{Real-Time Open-Vocabulary Semantic Mapping --}CrossMaps performs online semantic mapping by combining multi-scale vision–language embeddings with SLAM-based spatial alignment. Each incoming RGB frame is divided into tiles at multiple spatial scales (\Cref{fig:CLIP}), and each tile is encoded with a pretrained CLIP visual transformer (ViT-L/14, OpenAI weights) to obtain a semantic embedding aligned with natural language. These tile-level embeddings support open-vocabulary queries (e.g., “duck”) while avoiding the cost of dense, pixel-wise embeddings that require pre-segmentation. To reduce computation and increase robustness, we adopt tile-level aggregation instead of pixel-wise embeddings: this trades some spatial precision for higher throughput and resilience to camera motion. Using the corresponding depth image $D_t$ and viewpoint $T_t$, depth pixels are sampled and back-projected into 3D, transformed into the global frame, and associated with the fused embeddings of their corresponding coarse and fine image tiles. The resulting point-wise semantic observations are then accumulated in a top-down spatial grid map whose cells store semantic embeddings and statistics. Multi-view observations are fused via a top-down scheme that supports real-time mapping.

%CrossMaps operates as an online semantic mapping system that integrates multi-scale vision–language embeddings with SLAM-based spatial alignment. For each incoming RGB frame, the image is divided into tiles at multiple spatial scales~\Cref{fig:CLIP}, and each tile is encoded using a pretrained CLIP-based visual transformer (OpenAI L14) to produce a semantic embedding aligned with natural language representations. These embeddings capture high-level semantic information that allows map cells to later be queried with arbitrary text prompts, e.g., "duck". Unlike pixel-wise embedding approaches that rely on pre-segmentation for high precision, we propose to use tile-level aggregation, which substantially reduces computational cost and improves robustness to camera motion, though heterogeneous content within tiles can introduce noisy embeddings that are mitigated through filtering strategies. Using the corresponding depth image $D_t$ and rover viewpoint $T_t$, each tile is back-projected into 3D space and transformed into the global coordinate frame. The projected observations are accumulated in a spatial grid map where each cell stores semantic embeddings and associated statistics, integrating multi-view observations via a top-down fusion scheme. By trading a small loss in precision for significantly faster computation, this approach enables real-time semantic mapping while maintaining robustness, making it particularly suitable for mobile robotic platforms.

%--------------------------------------

\begin{algorithm}[t]
\footnotesize
\caption{CrossMaps Semantic Mapping Pipeline}
\label{alg:crossmaps}
\setlength{\parskip}{0pt}
\setlength{\itemsep}{0pt}
\begin{algorithmic}[1]

\REQUIRE RGB image $I_t$, depth map $D_t$, rover viewpoint $T_t$, confidence $\tau_c$ and coherence $\tau_h$ thresholds

\STATE Compute CLIP embeddings $e_i$ for all coarse and fine image tiles in Image $I_t$
\FOR{each sampled depth point $p$}    
\STATE {Back-project $p$ using $D_t$ and transform using $T_t$; retrieve corresponding coarse- and fine-tile embeddings; fuse them into a point embedding $e_p$; estimate geometric confidence $c^{geom}_{p}$; assign point observation to grid cell $\mathcal{C}$}
\ENDFOR

\FOR{each affected grid cell $\mathcal{C}$}
    \STATE Compute semantic consistency + temporal decay; update embedding, $c_{\mathcal{C}}$, $coh_{\mathcal{C}}$
\ENDFOR

\FOR{each STM cell $\mathcal{C}$}
    \IF{$C_{\mathcal{C}} > \tau_c \,\AND\, H_{\mathcal{C}} > \tau_h \,\AND\, \mathrm{viewpoint.diversity}$}
        \STATE Promote $\mathcal{C}$ to LTM
    \ENDIF
\ENDFOR

\STATE \textbf{Query:} encode + execute $q$; generate heatmap; select top navigation targets

\end{algorithmic}
\end{algorithm}

%%%%%%%%%%%%%%%%%%%%%
%\vspace{-3mm}
\vspace{-5mm}
%(Contribution 1)

\paragraph{Confidence-Aware STM Semantic Fusion --}After associating sampled 3D observations with fused tile-level semantic embeddings and assigning them to spatial grid cells, CrossMaps explicitly models perception uncertainty by assigning confidence values to observations during map fusion—see~\Cref{fig:pointConf}. Confidence is computed at both the \emph{point-level} and \emph{cell-level} using \emph{geometric}, \emph{semantic}, and \emph{temporal} cues. \textbf{Geometric confidence} captures viewing conditions such as sensor distance and visibility; distant observations are down-weighted via distance-based decay, while repeated observations from multiple viewpoints increase confidence. \textbf{Semantic confidence} measures agreement between new embeddings and the current cell evidence via cosine similarity in CLIP space, reinforcing consistent observations and down-weighting inconsistent ones, while a coherence measure assesses agreement among fused observations. \textbf{Temporal confidence} applies time-based decay so that confidence in rarely observed regions decreases, enabling the STM to adapt to changes while the LTM remains persistent.

These confidence components are combined into confidence-weighted updates during semantic cell fusion so that reliable observations influence the map more strongly. To fuse multiple observations in the same grid cell, the STM performs confidence-weighted semantic fusion: each cell $\mathcal{C}$ maintains a semantic accumulator $S_{\mathcal{C}}$ and a confidence accumulator $W_{\mathcal{C}}$, where $S_{\mathcal{C}}$ stores the sum of confidence-weighted embeddings and $W_{\mathcal{C}}$ stores the accumulated confidence used for normalization—see~\Cref{fig:Cell}.

Before fusion, the confidence of a new observation $i$ is computed from the geometric, semantic, and temporal factors, and the stored cell confidence is decayed to reduce the influence of outdated observations. The incoming observation is then checked for semantic consistency with the current cell embedding using cosine similarity in CLIP space and softly down-weighted through a gating function if it deviates. The fusion step performs confidence-weighted aggregation of embeddings  $\mathbf{e}_i$ with confidence $c_i$ and is given by:
\( S_{\mathcal{C}} \gets S_{\mathcal{C}} + c_i \mathbf{e}_i, \quad
W_{\mathcal{C}} \gets \lambda W_{\mathcal{C}} + c_i,
\quad
\mathbf{e}_{\mathcal{C}}=\frac{S_{\mathcal{C}}}{\lVert S_{\mathcal{C}} \rVert} \), where $\lambda$ denotes the temporal decay factor and $\mathbf{e}_{\mathcal{C}}$ is the normalized semantic cell embedding. Afterwards, viewpoint coverage and semantic coherence statistics are updated to assess the reliability of the fused representation.

\paragraph{STM to LTM Promotion --}To maintain both adaptability and stability, CrossMaps generates a dual-memory semantic map representation consisting of STM and LTM. While the STM acts as a dynamic grid map that aggregates recent observations, the LTM represents semantic landmarks and other stable structures in the environment (e.g., furniture). When sufficient evidence accumulates for a cell in the STM, the pipeline evaluates whether the observation is reliable enough to be promoted to the LTM. Only cells that exhibit high confidence, strong semantic coherence, and support from multiple viewpoints are transferred to the LTM. CrossMaps therefore selects STM cells for promotion using three criteria—confidence, semantic coherence, and viewpoint diversity.

%\noindent\textbf{Confidence} -- A cell is considered for promotion if its accumulated confidence exceeds a predefined threshold ($\tau_c$ ). High-confidence observations are therefore transferred to the LTM, while low-confidence observations, typically caused by noise or unstable detections, remain in the STM and are eventually suppressed through temporal decay. If a new observation with sufficiently high confidence contradicts a previously stored LTM cell, the updated observation may replace the older entry, allowing the long-term map to adapt to environmental changes.

\noindent\textbf{Confidence} --
A cell is eligible for promotion when its accumulated confidence ($W_{\mathcal{C}}$) exceeds a threshold ($\tau_c$); high-confidence observations are moved to LTM, while low-confidence, noisy, or unstable detections remain in STM and decay over time. If a new high-confidence observation contradicts an existing LTM cell, it can replace the older entry, allowing the long-term map to adapt.

\noindent\textbf{Coherence} -- 
Beyond confidence, we measure how consistently observations within a cell agree in embedding space. Let $\mathbf{e}_i$ be the normalized embeddings fused into a cell; coherence is the norm of their mean  \(\mathrm{coh}_{\mathcal{C}} =
\left\|
\frac{1}{n}\sum_{i=1}^{n}\mathbf{e}_i
\right\|\), which estimates angular variance in embedding space. Well-aligned embeddings yield a large norm and coherence near $1$, while conflicting embeddings cancel out and drive coherence toward $0$, so only cells with coherence above a threshold ($\tau_h$) are promoted.

%\noindent\textbf{Coherence.}In addition to confidence, we measure the semantic consistency of the observations within a cell. Coherence captures how well the accumulated embeddings agree in the embedding space. Let $\mathbf{e}_i$ denote the normalized embeddings fused into a cell. The coherence is computed as the norm of the mean embedding \(\mathrm{coh}_{\mathcal{C}} =
%\left\|
%\frac{1}{n}\sum_{i=1}^{n}\mathbf{e}_i
%\right\|\) which corresponds to an angular variance estimate in the embedding space. If the embeddings are well aligned, the mean vector has a large norm and the coherence approaches $1$. Conversely, if the embeddings point in different directions, they cancel each other out and the coherence approaches $0$. This measure therefore suppresses noisy cells and promotes semantically consistent observations, i.e., above a predefined threshold ($\tau_h$).

\noindent\textbf{Viewpoint Diversity} --
Finally, CrossMaps requires confirmation from multiple viewpoints. For each cell, the relative rover–cell angle is discretized into viewpoint bins and stored as a bitmask (e.g., $00011011$) that records viewing directions. Cells observed from several viewpoints are treated as more reliable, are more likely to be promoted to LTM, and can receive initial boosted confidence in STM.

\paragraph{Natural-Language Semantic Querying --}The CrossMaps semantic map supports open-vocabulary queries by embedding natural language into the same space as map cells. Given a query $q$, we compute its CLIP ViT-L/14 text embedding \(\mathbf{e}_{q} = \text{CLIP}_{\text{text}(q)}\) and for each cell with embedding \(\mathbf{e}_{\mathcal{C}}\), we evaluate the \emph{cosine similarity} (as semantic matching score) 
\(
cos(\mathbf{e}_{q}, \mathbf{e}_{\mathcal{C}}) = \frac{\mathbf{e}_{q} \cdot \mathbf{e}_{\mathcal{C}}}{\|\mathbf{e}_{q}\| \, \|\mathbf{e}_{\mathcal{C}}\|}
\label{eq:cosin}
\). To obtain a spatial heatmap $H$, we weight this similarity by the cell coherence \(coh_{\mathcal{C}}\) as 
\(
H_{\mathcal{C}} = s_{\mathcal{C}} \cdot coh_{\mathcal{C}}
\label{eq:heatMap}
\), so that semantically consistent regions remain visible even at lower confidence. For the top-ranked cell, its raw confidence \(c_{\mathcal{C}}\) influences ranking, yielding a natural-language-driven spatial index where high-confidence cells act as landmarks while potentially useful low-confidence regions are still highlighted.

%\paragraph{Natural-Language Semantic Querying --}The resulting semantic map in CrossMaps supports open-vocabulary queries. Given a natural language query, we first embed it into the same space as the semantic map cells using the CLIP ViT-B/32 text encoder, producing 
%\(\mathbf{e}_{q} = \text{CLIP}_{\text{text}(q)}\). For each cell with embedding \(\mathbf{e}_{\mathcal{C}}\), we compute the \emph{cosine similarity} 
%\(
%cos(\mathbf{e}_{q}, \mathbf{e}_{\mathcal{C}}) = \frac{\mathbf{e}_{q} \cdot \mathbf{e}_{\mathcal{C}}}{\|\mathbf{e}_{q}\| \, \|\mathbf{e}_{\mathcal{C}}\|}
%\label{eq:cosin}
%\) which measures how well the cell matches the query. To generate a heatmap $H$ over the spatial grid, we weight this similarity by the cell's coherence factor \(coh_{\mathcal{C}}\), preserving areas that are interesting even if their confidence is low:
%\(
%H_{\mathcal{C}} = s_{\mathcal{C}} \cdot coh_{\mathcal{C}}
%\label{eq:heatMap}
%\). 
%For the \emph{top-ranked cell}, its raw confidence \(c_{\mathcal{C}}\) is incorporated directly to influence ranking. This approach produces a natural-language-driven spatial index: the heatmaps highlight relevant areas across the map, allowing high-confidence cells to serve as reliable semantic landmarks while still keeping potentially informative low-confidence regions visible.

%----------------------------------------------------
\section{Implementation and Proof-of-Concept Demonstration}

%To validate the framework, we implemented CrossMaps as a real-time perception and semantic mapping pipeline on a Waveshare UGV Jetson Orin rover, %\footnote{\url{https://www.waveshare.com/ugv-rover-ros2-kit.htm}},
%which provides a compact, GPU-equipped platform suitable for onboard inference and real-time navigation. %The system processes RGB-D observations, extracts vision–language embeddings, and incrementally constructs a confidence-aware semantic map that supports open-vocabulary natural-language queries.

To validate the framework, we implemented CrossMaps as a real-time perception and semantic mapping pipeline on a Waveshare UGV Jetson Orin rover (a compact GPU-equipped platform).% suitable for onboard inference and real-time navigation.

\noindent\textbf{Pipeline Implementation} --
%CrossMaps is implemented as a modular ROS~2 node that operates alongside a SLAM-based localization pipeline. The node subscribes to RGB images, depth images, camera calibration data, and pose estimates through standard ROS~2 topics. CrossMaps maintains its internal semantic representation in the odometry frame, while published semantic maps are transformed into the global map frame via ROS~2 TF transforms. This separation allows the system to remain stable under SLAM updates such as loop closures, since semantic observations are stored in a consistent internal frame and re-projected into the corrected global frame during publication. Semantic features are extracted using a pretrained CLIP ViT-L/14 model, providing a balance between semantic expressiveness and computational efficiency for onboard inference. Incoming RGB frames are divided into multi-scale tiles, and the corresponding embeddings are projected into 3D using depth measurements and SLAM pose estimates. These observations are integrated into the semantic grid map described in~\Cref{sec:approach}. A set of configurable parameters controls mapping behavior, including grid resolution, STM decay rate, confidence thresholds, and promotion rules for promoting cells from STM to LTM. The code and replication package are publicly available~\cite{crossmaps2026}
CrossMaps is implemented as a modular ROS~2 node operating alongside a SLAM-based localization pipeline. The node subscribes to RGB and depth images, camera calibration data, and pose estimates via standard ROS~2 topics. Semantic maps are maintained in the odometry frame and transformed to the global map frame through ROS~2 TF, ensuring stability under SLAM updates such as loop closures. Semantic features are extracted using a pretrained CLIP ViT-L/14 model, balancing expressiveness and computational efficiency for onboard inference. RGB frames are divided into multi-scale tiles whose embeddings are projected into 3D using depth measurements and pose estimates. These observations are integrated into the semantic grid map described in~\Cref{sec:approach}. Configurable parameters control mapping behavior, including grid resolution, STM decay rate, confidence thresholds, and STM-to-LTM promotion rules. The code and replication package are publicly available~\cite{crossmaps2026}.

\noindent\textbf{Platform Deployment} -- 
Our prototype was deployed on a Waveshare UGV rover equipped with an RGB-D camera and an NVIDIA Jetson Orin GPU. The rover streams RGB-D observations and pose estimates via ROS~2 to an external computer running CrossMaps, while RTAB-Map provides localization for global alignment. CrossMaps processes these inputs to build semantic maps and publishes embeddings, confidence grids, and heatmaps via ROS~2 topics, enabling visualization in RViz.

\begin{figure}[h]
  \centering
  \begin{subfigure}[t]{0.2\linewidth}
    \centering
    \includegraphics[width=\linewidth]{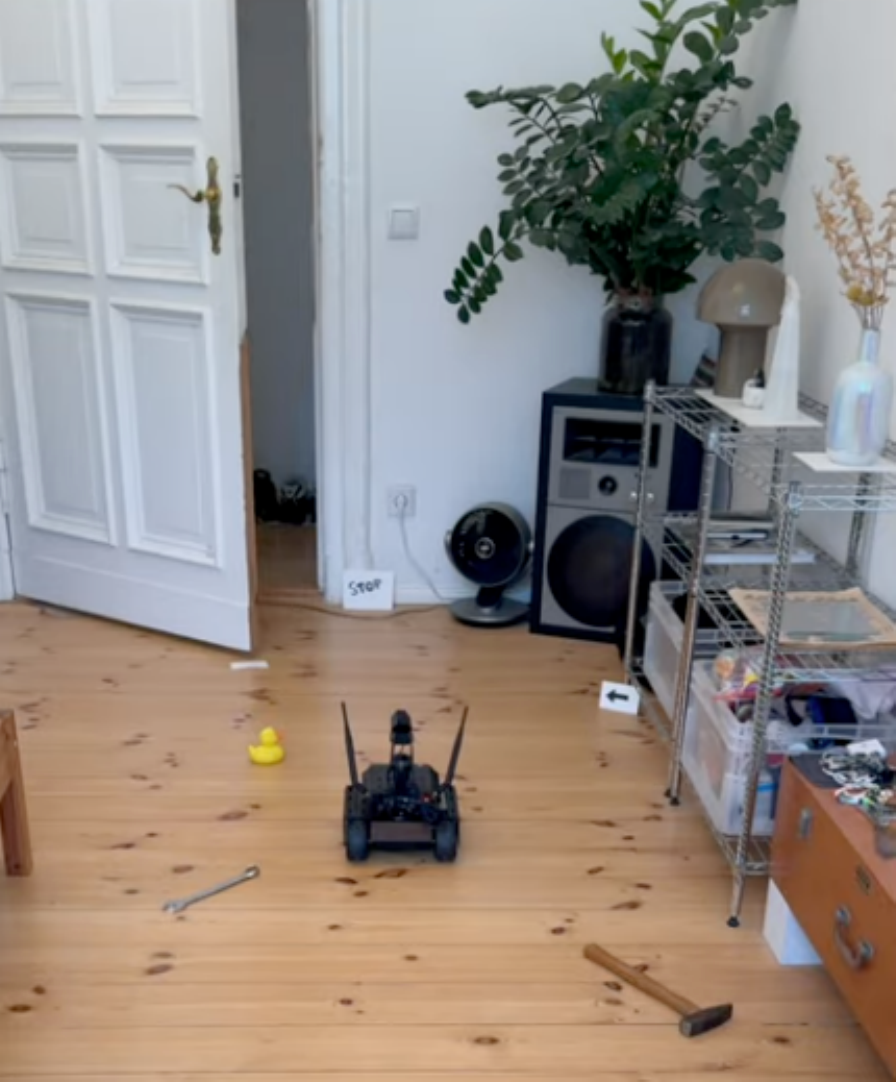}
    \caption{Rover in the room}
    \label{fig:room}
  \end{subfigure}
   \hfill
  \begin{subfigure}[t]{0.3\linewidth}
    \centering
    \includegraphics[width=\linewidth]{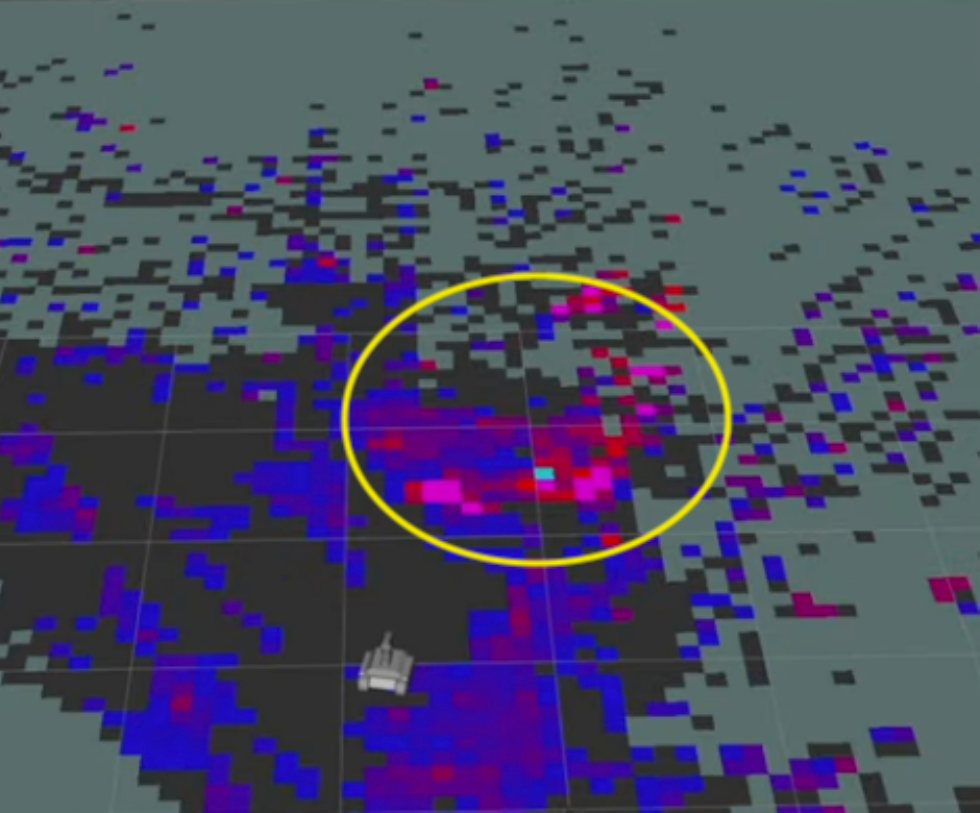}
    \caption{STM for query "plant" }
    \label{fig:plantmap}
  \end{subfigure}
  \hfill
  \begin{subfigure}[t]{0.25\linewidth}
    \centering
    \includegraphics[width=\linewidth]{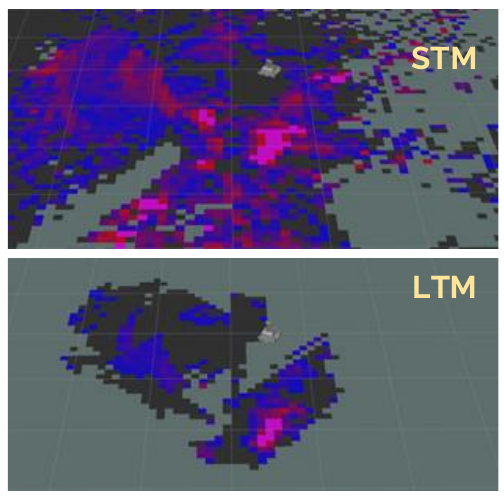}
    \caption{Query "hammer"}
    \label{fig:LTM}
  \end{subfigure}

  \caption{STM and LTM heatmaps for different natural language queries in CrossMaps}\label{fig:heatmaps}
\end{figure}

\noindent\textbf{Semantic Query Demonstration} -- 
CrossMaps supports open-vocabulary semantic queries at runtime. A natural-language query is encoded using the CLIP text encoder and compared with stored cell embeddings using cosine similarity. The resulting similarity scores are visualized as spatial heatmaps over the semantic grid map. \Cref{fig:heatmaps} illustrates representative heatmaps generated for example queries such as "plant" and "hammer". These heatmaps highlight map regions that match the query while suppressing inconsistent observations through the STM update and confidence mechanisms. As in \Cref{fig:heatmaps}, for the query "hammer", the STM tends to be noisy, while LTM shows more sparse and reliable suggestions. A demonstration of the rover exploration is available at \url{https://youtu.be/vMQndtoBYTU}.

\vspace{-3mm}
\vspace{-3mm}

\section{Conclusion, Future Work and Declaration on Generative AI}\label{sec:conclusionFW}

CrossMaps shows that combining open-vocabulary perception with confidence-aware fusion and a dual-
memory representation enables rovers to build language-queryable semantic 
maps while mitigating perceptual noise. The STM absorbs uncertain observations, while the LTM retains only stable semantic structures supported by sufficient confidence and coherence. Our proof-of-concept demonstrates real-time operation with natural-language queries, and future work will integrate segmentation-based feature extraction~\cite{Kirillov2023SAM} and extend CrossMaps toward multi-agent sharing of causal knowledge~\cite{korte2025causal} and neuro-symbolic reasoning~\cite{adriano2025neuro, adriano2024principled, Rehan2026NeuroSymbolicCausal} to tackle safety-critical rover navigation scenarios. Generative AI tools were used solely for language editing. The authors are responsible for the content.
%\printbibliography

\bibliography{references}

\end{document}